\documentclass[runningheads]{llncs}

\usepackage{eccv}

\usepackage{eccvabbrv}

\usepackage{graphicx}
\usepackage{booktabs}

\usepackage[accsupp]{axessibility}  
\usepackage{soul}

\usepackage{hyperref}

\usepackage{orcidlink}

\begin{document}

\title{SkelMo: Universal Skeletal Motion Generation for 3D Rigged Shapes} 

\titlerunning{SkelMo}

\author{Ye Tao\orcidlink{0009-0004-5923-0692} \and
Yuxin Yao\textsuperscript{*}\orcidlink{0000-0002-5410-0782} \and
Kendong Liu\orcidlink{0009-0003-6063-9674} \and
Dapeng Wu\orcidlink{0000-0003-1755-0183} \and
Junhui Hou\thanks{Corresponding authors. Email: \{yuxinyao,jh.hou\}@cityu.edu.hk}\orcidlink{0000-0003-3431-2021}
}

\authorrunning{Y. Tao et al.}

\institute{Department of Computer Science, \\City University of Hong Kong, \\Hong Kong SAR, China}

\maketitle
\vspace{-0.8cm}
\begin{figure}[h]
    \centering
    \includegraphics[width=\linewidth]{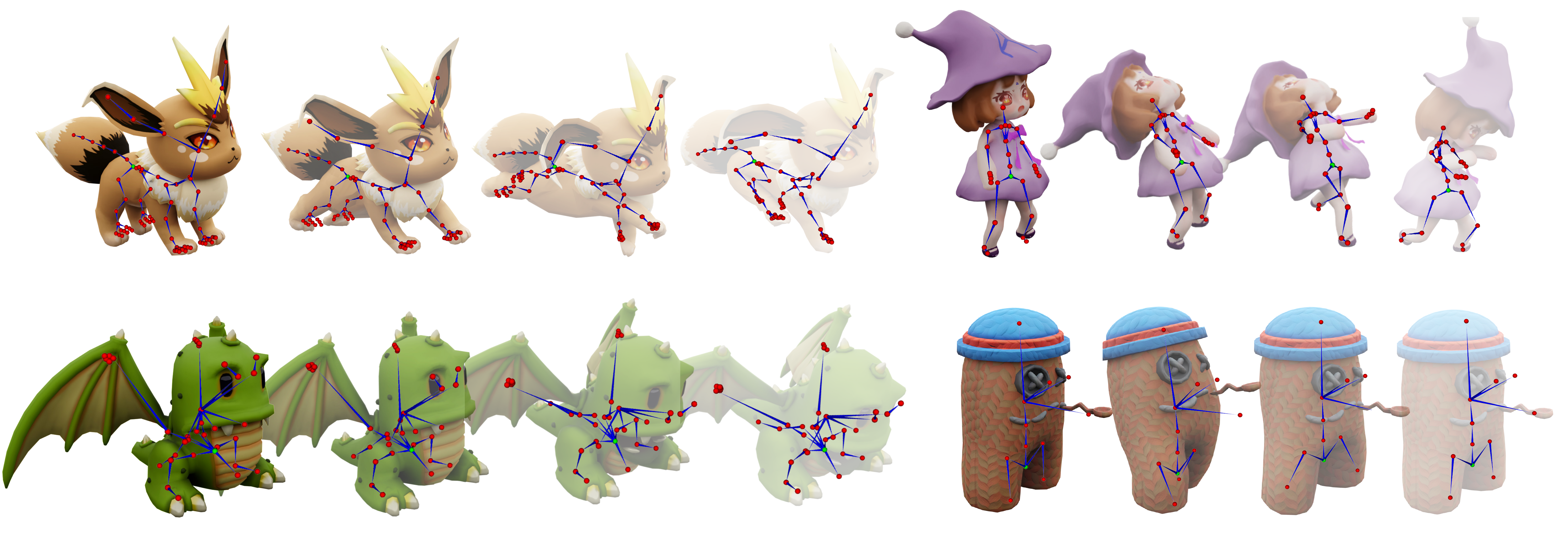} \vspace{-0.4cm}
    \caption{SkelMo utilizes topology agnostic diffusion to generate skeletal animation according to the given skeleton and driving video.
  }
  \vspace{-24px}
  \label{fig:teaser}
\end{figure}
\begin{abstract}
  Motion generation for rigged shapes is vital for scalable 4D asset production. However, template-based methods are limited by specific topologies and fail to generalize across diverse morphologies. Conversely, per-case optimization is computationally expensive, susceptible to local optima, and highly sensitive to viewpoint-induced ambiguities. In this paper, we present SkelMo, a diffusion-based framework designed for category-agnostic skeletal animation generation from 2D video guidance. To overcome the scarcity of high-quality training data, we have curated a large-scale dynamic dataset comprising approximately 20,000 diverse 3D animations, each featuring complete textures, skeletal rigging, and a wide array of comprehensive animation sequences. To bridge the kinematic gap between 2D visual motion cues and heterogeneous 3D skeletal structures, we propose a structural-semantic injection mechanism. Our model integrates texture and semantic attributes directly into skeletal joint representations. This allows it to map perceived visual dynamics to specific joint hierarchies and their functional roles.  This enables SkelMo to synthesize high-fidelity animations that maintain anatomical consistency across a vast range of unseen categories, from existing biological species to fantastical beings. Extensive experiments demonstrate that our approach significantly outperforms existing methods, setting a new state-of-the-art benchmark for robust and efficient 4D asset generation. Project Page: \url{https://research.davytao.me/skelmo/}.
  \keywords{Motion Generation \and Skeletal Animation \and Diffusion Model}
\end{abstract}

\section{Introduction}
\label{sec:intro}

Generating high-quality motion for rigged objects is a fundamental challenge in computer graphics and computer vision.
In the digital realm, this technology addresses a critical bottleneck in 4D asset creation by reducing the reliance on manual animation. 
As a result, it has the potential to transform immersive entertainment, including video games, virtual reality (VR), and digital avatars. 
Beyond virtual environments, robust motion generation also serves as a crucial bridge for embodied AI, enabling physical robots to execute realistic and physically plausible movements in the real world.

Traditionally, realistic motion is generated through two primary approaches. The first relies on expensive capture systems, such as calibrated multi-camera setups or IMU devices, which require specific environments and are often restricted to certain categories like human bodies. The second approach involves manual animation by professional artists, which is labor-intensive and difficult to scale for diverse virtual characters.
Recently, some frameworks \cite{yi2024generating,li2025morph,xiao2025motionstreamer,sun2024ponymation,zhang2025motionanything,chu2023command} have been proposed to automatically synthesize motion from various input modalities, such as text or video. However, due to limited diversity in training data, these methods are primarily restricted to human bodies, quadrupeds, or simple articulated shapes like drawers. 
AnyTop~\cite{gat2025anytop} introduces an architecture for universal motion generation. However, in the absence of fine-grained conditional control and due to the limited scale of training data (fewer than 100 objects), its practicality remains inherently constrained.
With the rapid advancement of video generative models~\cite{yang2024cogvideox,wan2025}, synthesizing motion from monocular videos has emerged as a promising direction. A straightforward approach, as seen in~\cite{Puppeteer,xie2025animamimic}, involves optimizing skeletal poses to align the deformed shape with the video frames. Nevertheless, such methods are often sensitive to rapid movements and rely heavily on external priors like depth estimation or point tracking. Furthermore, the requirement for case-specific optimization leads to significant computational overhead.

In this paper, we propose SkelMo, a diffusion-based framework for category-agnostic skeletal animation generation.
A major hurdle in training such a generative model is that high-quality 3D models with both skeletal rigging and diverse animations are scarce in existing open-source datasets.
In response to this data scarcity, we curate a large-scale dynamic dataset consisting of approximately 20,000 carefully selected 3D animations.
Each model is equipped with high-fidelity textures, complete skeletal rigging, and a wide array of skeletal animation sequences.
The dataset is meticulously preprocessed with rest-pose normalization and orientation alignment to ensure cross-category consistency.
Building on this foundation, our video-conditioned model generates skeletal animation parameters for input 3D models, with the ability to generalize across diverse shape types.
To achieve true category-agnostic generalization, a key challenge lies in aligning 2D visual motion cues with the diverse 3D skeletal topologies of heterogeneous objects.
To address this, we encode the textural and semantic information of each 3D model into its skeletal joint representations, allowing the model to effectively correlate visual motion patterns from the video with object-specific features and achieve better generalization.

In summary, our contributions are as follows.
\begin{itemize}
    \item We curate a high-quality dataset of approximately 20,000 3D animations featuring complete textures, skeletal rigging, and diverse animation sequences, all preprocessed for cross-category consistency.
    \item We propose SkelMo, a diffusion-based framework technically featured with a bidirectional video-skeleton fusion and a texture-semantic injection mechanism to enable precise alignment between visual motion cues and diverse 3D morphologies.
    \item We demonstrate that our approach pushes the performance of skeletal motion synthesis to a new height, setting a new state-of-the-art benchmark for generalization across unseen categories and complex motion sequences.
\end{itemize}

\section{Related Work}

\noindent \textbf{Motion Capture.}
Motion capture is traditionally categorized into two primary domains: human-centric and non-human domains. For human subjects, state-of-the-art methods predominantly focus on recovering the pose and shape of parametric body models from monocular or multi-view inputs (e.g., HMR~\cite{HMR}, Vibe~\cite{vibe}, TCMR~\cite{TCMR}). A dominant line of work regresses SMPL~\cite{loper2023smpl} or SMPL-X~\cite{pavlakos2019expressive} parameters from images or video clips, leveraging strong template priors to ensure anatomical plausibility and stable tracking.
These approaches span optimization-based fitting (e.g., SMPLify-X~\cite{pavlakos2019expressive}) and modern feed-forward one-stage regressors with ViT-style backbones for robust whole-body recovery (e.g., OSX~\cite{osx}, Aios~\cite{AiOS}).
To obtain physically meaningful trajectories, recent monocular systems incorporate SLAM/visual odometry and motion priors to infer world-grounded global motion (e.g., Pace~\cite{pace}, Tram~\cite{tram}), yet they still rely on a fixed human template and are not designed for arbitrary skeletons.
Non-human motion capture and reconstruction is often studied through model-free (e.g., CMR~\cite{CMR}, Lassie~\cite{Lassie}, Magicpony~\cite{magicpony}, and Lasr~\cite{lasr}) or model-based (e.g., SMAL~\cite{zuffi20173d}) animal pipelines.
Although effective within their target domains, these methods are typically species- or template-specific, which limits their ability to generalize to the diverse and often non-animal skeletal structures found in arbitrary animatable assets. 
Concurrently, MoCapAnything~\cite{gong2025mocapanything} introduces a monocular video-based motion prediction method for category-agnostic skeletons. However, its performance is constrained by the reliance on estimated meshes and limited training data.

\vspace{0.5em}
\noindent \textbf{4D Content Generation.}
4D generation, which focuses on dynamic 3D content synthesis, has recently attracted increasing attention.
Current 4D generation methods primarily differentiate themselves through various combinations of 3D representations, temporal modeling techniques, and algorithmic paradigms.
Early approaches (e.g., Sv4d~\cite{Sv4d}, Sv4d2.0~\cite{Sv4d2}, 4Diffusion~\cite{4diffusion} and DreamGaussian4D~\cite{DreamGaussian4D}) predominantly adopted an optimization-based framework to recover dynamic 3D models by lifting 2D video observations into the 3D space.
While these methods are category-agnostic, they are often computationally expensive and struggle to maintain spatio-temporal consistency across long sequences.
To address these efficiency bottlenecks, subsequent research (e.g., AnimateAnyMesh~\cite{wu2025animateanymesh}, DriveAnyMesh~\cite{shi2025drive}, and GVFD~\cite{zhang2025gaussian}) explored feed-forward networks to directly predict deformation fields or motion trajectories for temporal representation.
Although these methods offer significantly faster inference, they frequently suffer from geometric artifacts where rigid regions undergo unnatural non-rigid distortions.
Recently, as skeletal generative models (e.g., RigAnything~\cite{Riganything}, Magicarticulate~\cite{Magicarticulate}, Puppeteer~\cite{Puppeteer}, and Shapegen4d~\cite{Shapegen4d}) have matured, skeleton-based animation generation has emerged as a more principled alternative for achieving structured and plausible motion.
Despite its potential, research in this direction remains in its infancy, with AnyTop~\cite{gat2025anytop} being one of the few pioneering works to explore unconditional skeletal animation generation.

\section{Dataset Construction}
\label{sec:dataset}
\begin{figure}[t]
    \centering
     \includegraphics[width=\linewidth]{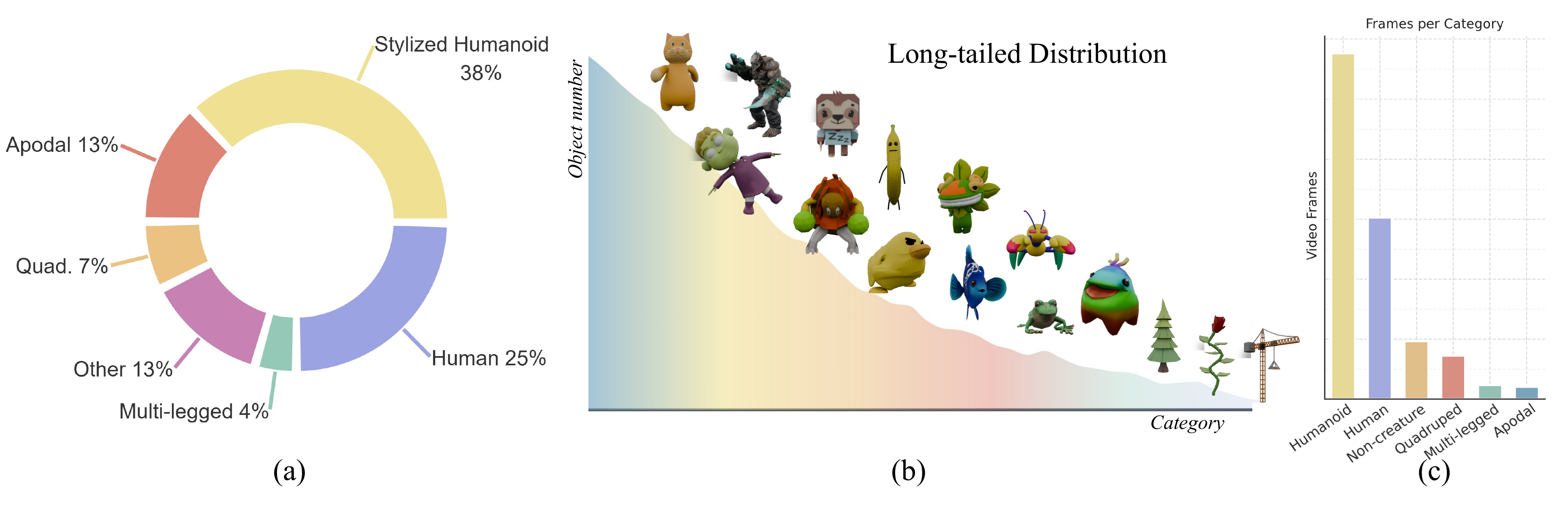}
     \vspace{-0.7cm}
    \caption{Dataset analysis and statistics. (a) Category proportions. (b) Object counts follow a long-tailed distribution across categories. (c) Video-frame totals are substantial overall. 
  }
  \vspace{-0.4cm}
  \label{fig:dataset}
\end{figure}
To empower SkelMo with the capability of animating arbitrary characters, we first establish a large-scale, category-diverse foundation.
Unlike previous datasets restricted to specific species, our collection encompasses a wide range of biological taxonomies, including humans, stylized humanoids, quadrupeds, bipeds, and multi-legged creatures, as statistically summarized in Fig.~\ref{fig:dataset}. We use LLM-assisted labeling for high-level category statistics. Overall, the category distribution exhibits a mild long-tail pattern while still providing broad coverage across diverse creature types.
The dataset comprises approximately 20,000 unique 3D animations and 
paired rendered videos and corresponding skeletal motion sequences, totaling more than 3.5 million frames of high-fidelity skeletal data. 
Each entry features a meticulously rigged skeleton sourced from Articulation-XL~\cite{Magicarticulate}, a curated subset of the Objaverse dataset~\cite{deitke2023objaverse, deitke2023objaversexl}.
This refined skeletal structure provides the necessary foundation for accurate and anatomically sound motion transfer.
To populate this dataset with diverse motions, 
we developed an automated pipeline that re-associates high-quality animation clips from the original Objaverse source and transfers them onto the refined skeletal structures using AutoRig-Pro~\cite{autorigpro}.
This process effectively recovers missing motion data while ensuring anatomical consistency.
Furthermore, we leverage the spatial reasoning capabilities of the Qwen large language model~\cite{bai2023qwen} to ensure that all models are oriented in a consistent and correct forward-facing  direction.
To support video-based conditioning, each animation sequence is rendered from a canonical front-view perspective, resulting in a large-scale collection of paired video-motion data.
This data diversity allows SkelMo to learn a generalized mapping between visual appearance and underlying articulation, thereby overcoming the limitations of template-specific priors such as SMPL~\cite{loper2023smpl} or SMAL~\cite{zuffi20173d}.

\section{Proposed Method}
\noindent\textbf{Overview.} 
Given a category-agnostic 3D mesh $\mathcal{M}=\{\mathcal{V}, \mathcal{F}\}$ with its associated skeleton 
$\mathcal{S}^0=\{\mathcal{P}^0, \mathcal{G}\}$,
where $\mathcal{V}$ and $\mathcal{F}$ denote the vertex positions and triangle faces, while $\mathcal{P}^0\in\mathbb{R}^{J\times 3}$ and $\mathcal{G}$ represent the joint positions and topological hierarchy respectively. Along with a monocular driving video $\mathcal{X}=\{I^l\}_{l=1}^{L}$, which can be synthesized by a generative model, our objective is to produce a sequence of posed skeletons $\{\mathcal{S}^{l}=(\mathcal{P}^l, \mathcal{G})\}_{l=1}^{L}$. This sequence should faithfully animate the mesh $\mathcal{M}$ according to the motion dynamics observed in $\mathcal{X}$.

\begin{figure}[t]
    \centering
    \includegraphics[width=\linewidth]{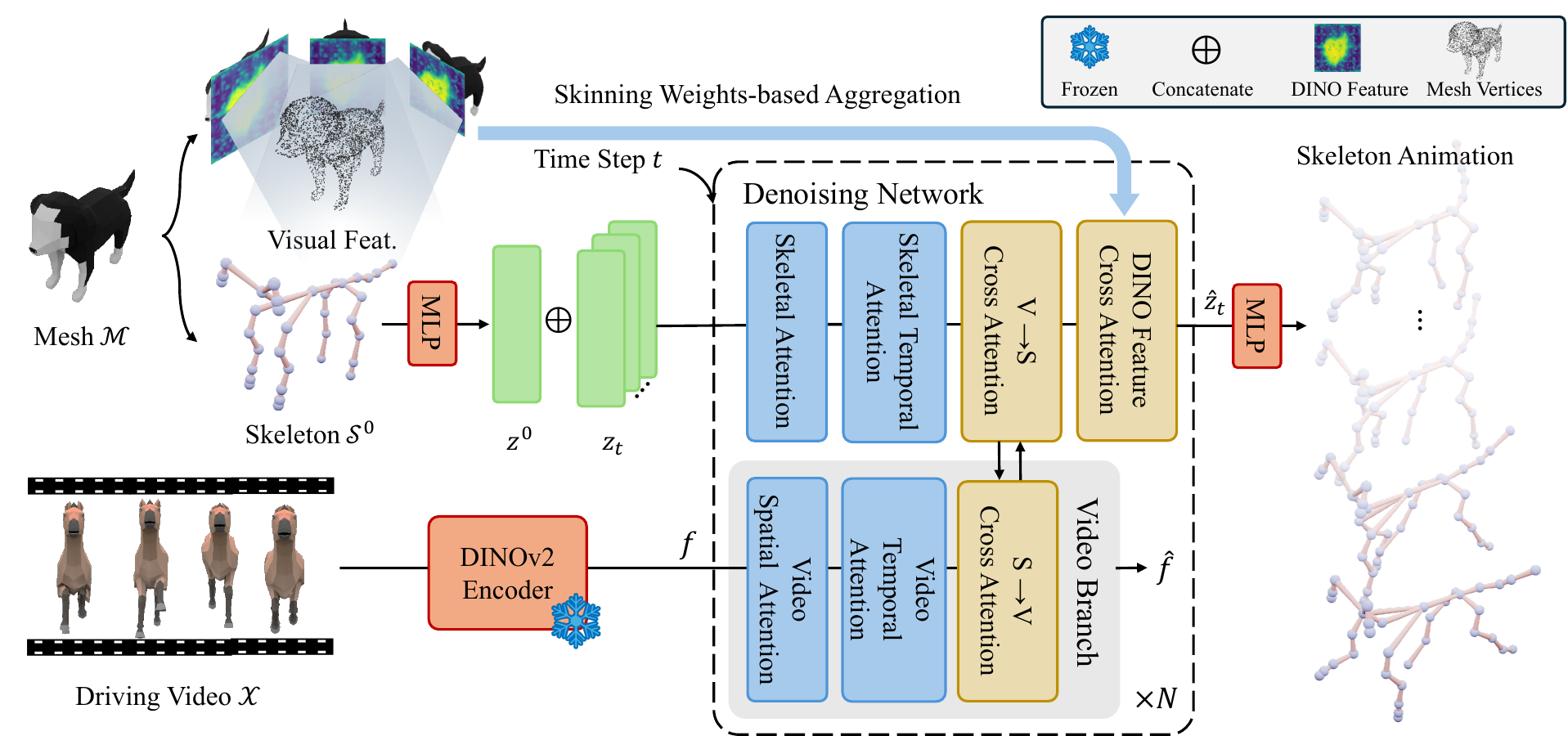}
    \caption{\textbf{Overview of SkelMo Pipeline.} Given a mesh $\mathcal{M}$ with its skeleton $\mathcal{S}^0$ and driving video $\mathcal{X}$, our approach follows a structured pipeline: At each sampling step $t$, the driving video $\mathcal{X}$ is encoded using DINOv2 Encoder, and fed into the video branch. Simultaneously, the initial skeleton $\mathcal{S}^0$ is projected into the latent space as $z^{0}$ via a linear layer. This condition is then concatenated with the current noisy latents $z_t$, serving as a structural prior. Both noisy latent $z_t$ and its corresponding video feature $f$ are processed through the denoising network consisting of $M$ repeated blocks, and a bidirectional video-skeleton fusion module then integrates these features. To incorporate geometric constraints, multi-view visual features are extracted from the static mesh $\mathcal{M}$, projected onto the vertices. We apply skinning weights-based aggregation to the mesh vertices to extract representative visual features. These features are then injected into the fused representation via cross attention, where skinning weights serve as a masking mechanism to ensure anatomically plausible feature alignment. For the last step, the resulting denoised latent $\hat{z}_0$ is passed through a linear layer to produce the final skeletal animation $\{\mathcal{P}^l\in\mathbb{R}^{J\times 3}\}_{l=1}^{L}$.
  }
   \vspace{-0.7cm}
  \label{fig:overview}
\end{figure}

To achieve this, we develop a topology-agnostic diffusion model as our backbone architecture (Sec.~\ref{sec:anytop}).
Building upon this foundation, we propose SkelMo, a diffusion-based framework illustrated in Fig.~\ref{fig:overview}  that effectively bridges visual motion cues with structured skeletal hierarchies.
SkelMo consists of two core technical modules.
First, to bridge the gap between 2D visual semantics and 3D structural joints, we develop a skinning-aware texture-semantic injection module (Sec.~\ref{sec:injection}). 
Second, we implement a bidirectional motion-topology alignment mechanism within a diffusion transformer (Sec.~\ref{sec:bidirection}) to reconcile driving video cues with the target skeletal hierarchy. 
This systematic approach ensures that the generated motions are both high-fidelity to the source video and anatomically consistent with the target morphology.

\subsection{Topology Agnostic Diffusion Model}
\label{sec:anytop}
To animate characters with arbitrary skeletal configurations, we require a generative backbone that is independent of specific joint templates or hierarchical constraints. The primary challenge in such a topology-agnostic setting is maintaining spatial consistency between the driving 2D pixels and the 3D joints without a predefined structural prior.

\vspace{0.5em}
\noindent\textbf{Motion Representation.} In contrast to traditional methods that rely on rotation-based relative representations, we adopt global Cartesian coordinates $\mathcal{P} \in \mathbb{R}^{L \times J \times 3}$ as our primary motion descriptor. While relative rotations excel in local joint smoothness, they introduce complex, non-linear mappings during 2D-to-3D lifting, often leading to spatial misalignment in video-conditioned tasks. By directly diffusing in the $xyz$ coordinates space, we establish a more intuitive geometric correspondence with the driving video $\mathcal{X}$, which stabilizes the denoising process and ensures better global trajectory alignment.

\vspace{0.5em}
\noindent\textbf{Structural Conditioning.} To ensure that the generated coordinates respect the target morphology, we incorporate the rest-pose skeleton $\mathcal{S}^0$ as a persistent structural prior. Specifically, the joint positions $\mathcal{P}^0$ are projected into a latent space $z^0 \in \mathbb{R}^{J \times C}$ and concatenated with the noisy latent $z_t$ along the temporal axis at each diffusion step $t$. This conditioning allows the model to perceive the character's bone lengths and proportions as a constant geometric anchor, preventing anatomical distortions during the iterative sampling process.

\vspace{0.5em}
\noindent\textbf{Topology-Aware Denoising.} Our denoising network consists of $M$ transformer blocks optimized for skeletal data. To capture dependencies across arbitrary hierarchies, we employ the Skeletal Attention mechanism~\cite{gat2025anytop}. This module modulates the self-attention weights using a tree-based distance matrix derived from the topology $\mathcal{G}$, ensuring that joints with direct biological relationships (e.g., parent-child) maintain higher spatial correlation. The final denoised latent $\hat{z}_0$ is projected back to the coordinate space to produce the animation $\{\mathcal{P}^l\}_{l=1}^{L}$.

\subsection{Skinning-Aware Texture-Semantic Injection}
\label{sec:injection}
While the backbone captures geometric constraints, a purely structural representation lacks the semantic context of the shape's appearance.
To provide the model with a fine-grained understanding of the appearance of the object and its semantic structure, we design a specialized DINO-based feature injection mechanism.
Specifically, for a given 3D mesh $\mathcal{M}$, we render a set of static images $\{I_i^{0}\}^N_{i=1}$ from $N$ diverse viewpoints distributed around the object.
We utilize a pretrained DINOv2~\cite{oquab2023dinov2} model to extract dense visual descriptors $F_i = \Phi_{DINO}(I_i^0)$, where $F_i\in \mathbb{R}^{H\times W\times C}$ represents the spatial feature map for the $i$-th view.
To establish a surface-aligned representation, these 2D features are back-projected onto the visible vertices of the 3D mesh.
For each vertex $v\in\mathcal{V}$, its corresponding feature $f_v$ is computed by aggregating the projected descriptors from all viewpoints where $v$ is not occluded:
\begin{equation}
f_v = \text{Mean}(\{F_i(\pi_i(v)) \mid \text{Vis}(v, i) = 1\}),
\end{equation}
where $\pi_i$ denotes the projection function from the 3D space to the $i$-th image plane, and $\text{Vis}(v, i)$ is a binary visibility mask.
To effectively integrate high-dimensional surface semantics into the skeletal space, we leverage skinning weights as a geometric prior to guide the feature injection process.
Let $W=\{w_{j,v}\} \in \mathbb{R}^{J\times |\mathcal{V}|}$ denote the skinning weight matrix between $J$ skeletal joints and mesh vertices.
For the $j$-th joint, we normalize the weights such that the influence of all vertices sums up to unity:
\begin{equation}
    \hat{w}_{j,v} = \frac{w_{j,v}}{\sum_{v\in\mathcal{V}} w_{j,v}}, ~~~~\text{s.t.} \sum_{v\in\mathcal{V}} {\hat{w}_{j,v}} = 1.
\end{equation}
This normalization ensures that each joint acts as a localized aggregator for the visual semantics of its associated body parts.
The aggregated joint feature $\widetilde{F}_{\mathcal{S}} \in \mathbb{R}^{J\times C}$ is then computed through a direct linear projection:
\begin{equation}
    \widetilde{F}_{\mathcal{S}} =\hat{W}F_\mathcal{V},
\end{equation}
where $F_{\mathcal{V}}\in \mathbb{R}^{|\mathcal{V}|\times C}$ denotes the collected vertex features $\{f_v\}$.
Each row of $\widetilde{F}_{\mathcal{S}}$ encodes the semantic profile of a corresponding joint, distilled from its physically associated surface regions.
Once the joint-specific features $\widetilde{F}_{\mathcal{S}}$ are computed, they are injected into the motion synthesis network.
We employ a cross-attention mechanism where the queries $Q$ are the hidden states of the skeletal joints within the network, while the keys $K$ and values $V$ are derived from the pre-computed $\widetilde{F}_{\mathcal{S}}$.

\subsection{Bidirectional Video-Skeleton Fusion}
\label{sec:bidirection}
Building upon the static geometry and semantic features derived above, we now introduce how SkelMo synchronizes these priors with dynamic motion cues from the driving video.
This stage begins with a spatial-temporal video encoder to extract motion cues, followed by a bidirectional cross-modal interaction between motion cues and skeletal topology.

\vspace{0.5em}
\noindent\textbf{Video Encoder}. Given a driving video $\mathcal{X}$ consisting of $L$ frames, we first extract a sequence of raw visual descriptors $F_{\mathcal{X}} = \{f_{v,l}\}^L_{l=1}$ using a pre-trained DINOv2 backbone.
To capture global motion context and inter-frame dependencies within the model, we employ an integrated video encoder consisting of successive spatial and temporal self-attention layers.
In this stage, spatial attention captures the instantaneous pose semantics in each frame, while temporal attention encodes the dynamic trajectory across the sequence, yielding a motion-aware video representation $F^{global}_{\mathcal{X}}$.

\vspace{0.5em}
\noindent\textbf{Symmetrical Cross-modal Interaction}. The core of our fusion module is a bidirectional cross-attention mechanism that treats the skeletal features 
$\widetilde{F}_{\mathcal{S}}$ and refined video features $F^{global}_{\mathcal{X}}$ as dual sequences.
This symmetrical interaction is formalized through two parallel paths.
In the first path, the skeletal joints query motion cues from the video frames to guide the synthesis:
\begin{equation}
    F_{\mathcal{S}\gets \mathcal{X}} = \text{Softmax}(\frac{Q_{\mathcal{S}}K_{\mathcal{X}}^T}{\sqrt d})V_{\mathcal{X}},
\end{equation}
where $Q_{\mathcal{S}}$ is derived from the skeletal embeddings and $(K_{\mathcal{X}}, V_{\mathcal{X}})$ are projected from the video context.
Simultaneously, the second path enables the video features to be conditioned on the target skeletal topology:
\begin{equation}
    F_{\mathcal{X}\gets \mathcal{S}}=\text{Softmax}(\frac{Q_{\mathcal{X}}K_{\mathcal{S}}^T}{\sqrt d})V_{\mathcal{S}},
\end{equation}
where $d$ denotes the scaling dimension of the query and key features.
This bidirectional alignment ensures that the motion cues are not merely broadcast to the skeleton but are filtered and refined based on the specific joint hierarchy of $\mathcal{S}$. 
The fused features from both paths are then integrated into a unified spatio-temporal representation, which is fed into the subsequent diffusion denoising blocks.
This symmetrical interaction ensures that the generated animation maintains high fidelity to the source video while strictly adhering to the structural constraints of the target skeleton.

\subsection{Training Objective}
Our model is trained as a conditional diffusion model that generates 3D joint sequences. 
Following the standard diffusion formulation, we employ the simplified training objective.
The training process aims to predict the noise $\epsilon$ added to the ground-truth joint coordinates. 
Given the noise-scaled joint positions 
$z_t$ at diffusion step $t$, the loss function is defined as:
\begin{equation}
    L_{\text{diff}} = \mathbb{E}_{z_0, \epsilon, t} \left[ \| \epsilon - \epsilon_\theta(z_t, t, C) \|^2 \right],
\end{equation}
where $z_0$ represents the ground-truth 3D joint coordinates, $\epsilon \sim \mathcal{N}(0, \mathbf{I})$ is the Gaussian noise added at diffusion step $t$, and $\epsilon_\theta$ is our network that predicts the noise given the noisy sample $z_t$, the timestep $t$, and the video conditioning $C$.

\section{Experiments}
\subsection{Experiment Settings}

\noindent\textbf{Training Details.} We trained our model from scratch.
The model was trained for 80,000 steps with a batch size of 16.
We used AdamW with a learning rate starting with $1\times 10^{-4}$ to optimize the model.
We implemented a step decay schedule where the learning rate is multiplied by 0.99 every 10,000 steps.
The experiments were conducted on 4 NVIDIA RTX 5880-48G GPUs.

\vspace{0.5em}
\noindent\textbf{Post-processing.} To ensure kinematic consistency, we applied an IK-based refinement to the raw generated joints. Specifically, we projected the synthesized coordinates onto the predefined skeleton $\mathcal{S}^0$ through 150 iterations of Inverse Kinematics (IK), effectively eliminating bone stretching while maintaining the rigid geometric constraints of the target character.

\vspace{0.5em}
\noindent\textbf{Benchmark.} To evaluate SkelMo, we selected a standardized subset of 100 randomly chosen models from our constructed dataset.
We rendered a video of no more than 40 frames from the front view. The object was positioned at the origin, with the camera placed at a distance of 2 meters for rendering.

\vspace{0.5em}
\noindent\textbf{Baselines.} We compare our model against several state-of-the-art (SOTA) methods to evaluate its performance in rigged shape motion generation.
Ideally, the most relevant work to ours is the concurrent MocapAnything~\cite{gong2025mocapanything}; however, as their source code and models are not publicly available, a direct comparison is not feasible. 
We therefore select the following representative methods as our baselines. 
First, we consider Puppeteer~\cite{Puppeteer}, which is primarily a skeleton generation framework; however, one of its key applications involves generating animations via an optimization process guided by optical flow and depth estimation priors. 
Second, we evaluate against Motion 3-to-4~\cite{chen2026motion} and ActionMesh~\cite{sabathier2026actionmesh}, which represent the SOTA in video-to-4D generation by focusing on the synthesis of dense vertex trajectories.
To bridge the architectural gap between these paradigms and ensure a fair comparison, we implement a standardized post-processing protocol (details in the \textit{Supplementary Material}). 
Specifically, we fit the vertex trajectories produced by Motion 3-to-4 and ActionMesh into skeletal animations, allowing for a unified evaluation of motion quality and structural integrity across all methods.

\vspace{0.5em}
\noindent\textbf{Metrics.} To quantify the accuracy of the generated skeletal motion, we employ two key metrics: Mean Per-Joint Position Error (MPJPE) for joint localization and Chamfer Distance (CD) for overall geometric fidelity. 
For baseline methods that directly output mesh sequences, we first extract the underlying skeletal animation via a fitting process to ensure a consistent and fair comparison. We further evaluate robustness under auto-rigged skeletons.

\subsection{Comparison on Motion Generation}

\begin{table}[t]
  \caption{Quantitative comparison with state-of-the-art methods. We evaluate our SkelMo against several baseline methods on our proposed test dataset. MPJPE (Mean Per Joint Position Error) and CD (Chamfer Distance) are reported to measure motion accuracy and surface fidelity, respectively. Bold values indicate the best performance.
  } 
  \label{tab:comparison}
  \centering
  \begin{tabular}{@{}l|c|c|c|cc@{}}
    \toprule
    Metrics& Motion 3-to-4& Puppeteer& ActionMesh& SkelMo\\
    & \cite{chen2026motion}& \cite{Puppeteer}& \cite{sabathier2026actionmesh}& (Ours)\\
    \midrule
    MPJPE$\downarrow$ & 0.230 & 0.161 & 0.197 & \textbf{0.054}\\
    CD$\downarrow$ & 0.475 & 0.155 & 0.418 & \textbf{0.086}\\
  \bottomrule
  \end{tabular}
\end{table}

\begin{figure}[t]
    \centering    \includegraphics[width=\linewidth]{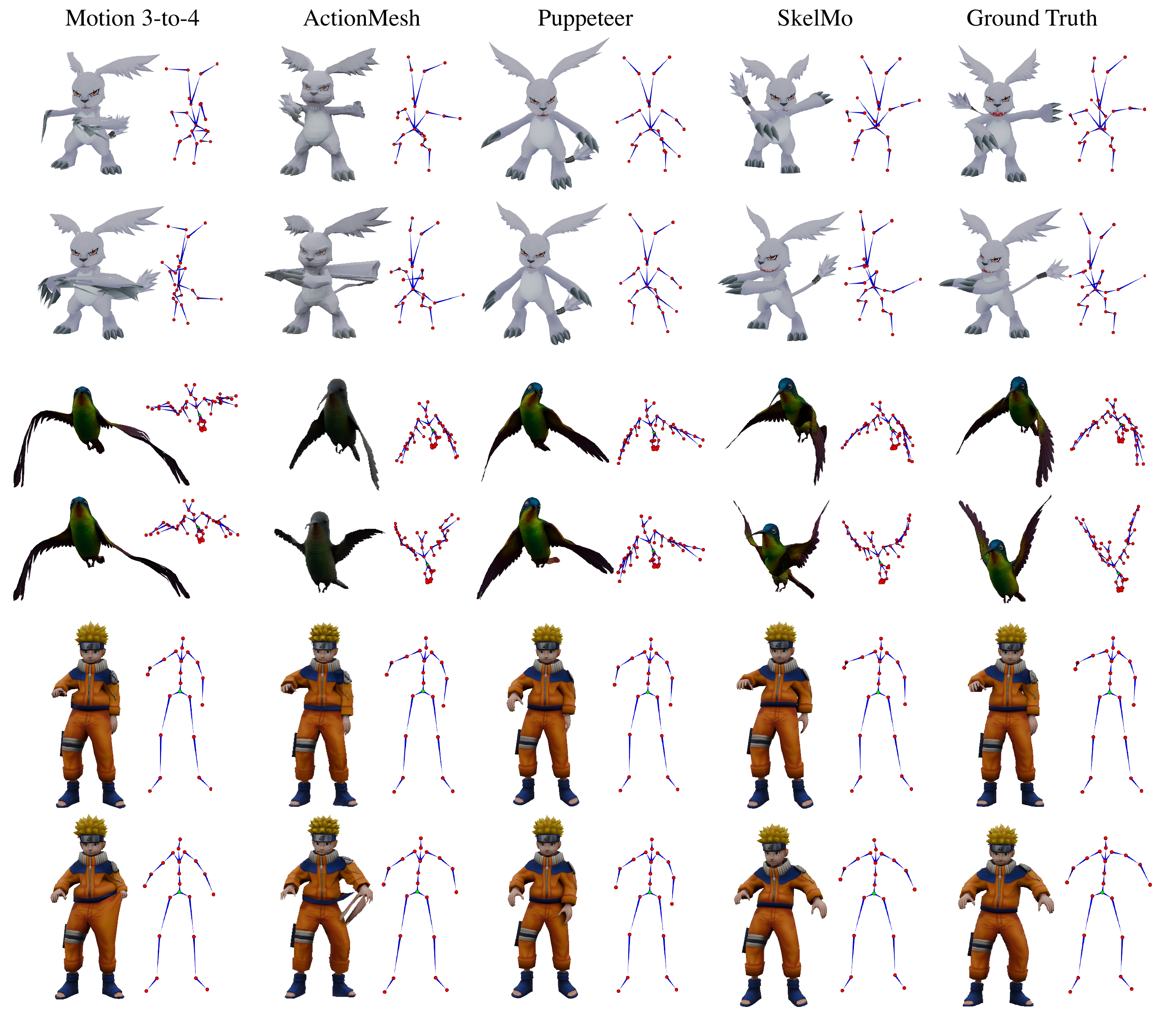}
    \caption{Visual comparison between our SkelMo and other state-of-the-art methods. For each example, the left side displays the rendered 3D mesh, while the right side shows its corresponding skeletal structure.
  }
  \label{fig:compare}
\end{figure}

In the quantitative evaluation, as shown in Table~\ref{tab:comparison}, SkelMo outperforms vertex-trajectory-based generative methods, such as Motion 3-to-4 and ActionMesh, across all metrics. 
Specifically, our method achieves a reduction in MPJPE, consistently outperforming Motion 3-to-4 and ActionMesh as shown in Table~\ref{tab:comparison}.
This lower error indicates a higher precision in capturing underlying motion dynamics. 
Furthermore, regarding the Chamfer Distance (CD), which measures geometric shape fidelity, we achieve a CD value that is lower than the competitive baselines, as detailed in Table~\ref{tab:comparison}.
These elevated CD scores suggest that the baseline methods struggle to preserve the original structural integrity of the object during motion.

These numerical findings are consistent with the visual comparisons presented in Fig.~\ref{fig:compare}. 
A primary limitation of pure vertex-trajectory generation is the susceptibility to non-rigid distortion. 
As observed in the case on the left, both Motion 3-to-4 and ActionMesh suffer from mesh adhesion and unnatural stretching artifacts. 
This phenomenon typically occurs because these frameworks independently predict vertex displacements, often lacking sufficient global constraints to maintain the spatial relationships between neighboring vertices. 
This leads to an elongated or "melted" appearance where the mesh topology is compromised. 
In contrast, our SkelMo maintains the rigid and articulated structures of the object more effectively, resulting in clean surfaces and stable geometry even during complex sequences. \textbf{The \textit{Supplementary Material} and \textit{Demo Video} provide more visual results.}
 
\subsection{Ablation Study}
\begin{table}[t]
  \caption{Ablation study on key components. We evaluate the contribution of each module to the overall performance. "w/o Inject." denotes the removal of texture-semantic injection; "w/o Bidir." refers to the absence of bidirectional video-skeleton fusion; and "Rel. Repr." represents the use of relative skeletal coordinates. The Full Model achieves the best performance across all metrics.
   } 
  \label{tab:ablation}
  \centering
  \begin{tabular}{@{}l|c|c|c|cc@{}}
    \toprule
    Setting & w/o Inject. & w/o Bidir. & Rel. Repr. & Full Model\\
    \midrule
    MPJPE$\downarrow$ & 0.234 & 0.118 & 0.285 & \textbf{0.054}\\
    CD$\downarrow$ & 0.235 & 0.147 & 0.340 & \textbf{0.086}\\
    
  \bottomrule
  \end{tabular}
\end{table}

\begin{figure}[t]
    \centering
    \includegraphics[width=\linewidth]{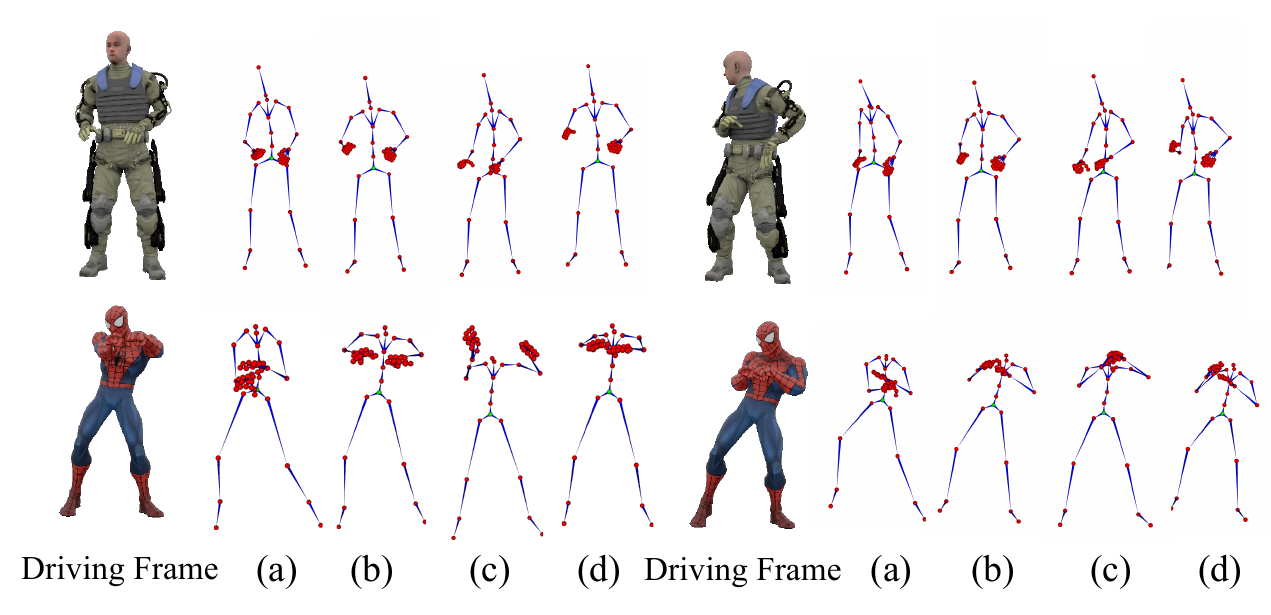}
    \caption{Visual comparison of different model variants.
From left to right: (a) Results without texture-semantic injection.
(b) Results without bidirectional video-skeleton fusion.
(c) Results using relative skeletal representations instead of global coordinates.
(d) Results of our full model.
Each column presents the generated skeletal poses for different input characters and driving frames.
  }
  \label{fig:ablation}
\end{figure}
In the ablation studies, all models are trained following the training procedures described previously.

\vspace{0.5em}
\noindent\textbf{Effect of Texture-Semantic Injection.} We investigate the impact of the texture-semantic injection module described in Sec.~\ref{sec:injection}. 
As shown in Table~\ref{tab:ablation}, removing the DINO Feature Cross Attention module leads to a measurable performance decline. 
As shown in Fig.~\ref{fig:ablation} (a), without the DINO-based semantic injection, the model struggles to identify the character's boundary and specific body parts. 
This leads to collapsed skeletal structures (e.g., the Spiderman example in the 2nd row), where the joints cluster together because the model lacks the semantic "priors" of what a skeleton should look like under that specific texture.

\vspace{0.5em}
\noindent\textbf{Effect of Bidirectional Video-Skeleton Fusion.} We evaluate the necessity of the bidirectional video-skeleton fusion module introduced in Sec.~\ref{sec:bidirection}. 
By ablating the $S \to \mathcal{X}$ and $\mathcal{X} \to S$ cross-attention mechanisms, we observe a consistent drop in performance, as shown in Table~\ref{tab:ablation}. 
As shown in Fig.~\ref{fig:ablation} (b), while the model can still capture the general motion, it fails to achieve fine-grained alignment with the driving frame’s subtle nuances (e.g., the exact angle of the elbows or the extension of the limbs).

\vspace{0.5em}
\noindent\textbf{Effect of Relative Skeletal Representations.} We examine the influence of skeletal representation methods as discussed in Sec.~\ref{sec:anytop}. 
Replacing global $xyz$ coordinates with relative representations leads to a performance degradation, as evidenced in Table~\ref{tab:ablation}. 
In Fig.~\ref{fig:ablation} (c), using relative coordinates instead of global ones leads to spatial disorientation. The model fails to maintain the correct global height and ground contact, often resulting in "floating" artifacts or inconsistent limb lengths.
This suggests that global spatial context provides more robust guidance for the model to capture absolute motion trajectories accurately.

\vspace{0.5em}
\noindent\textbf{Robustness to Auto-rigged Skeletons.} To evaluate robustness under imperfect rigging, we additionally test SkelMo with automatically generated skeletons (e.g., RigAnything~\cite{Riganything}) instead of ground-truth rigs. Since auto-rigged skeletons introduce topology and joint layout discrepancies, skeletal-space metrics become inconsistent; thus, we report vertex-space metrics instead. As shown in Table~\ref{tab:autorig}, SkelMo maintains strong performance under auto-rigged skeletons with only minor degradation.

\begin{table}[h]
\caption{Performance comparison under ground-truth (GT) and automatically generated (auto-rigged) skeletons. SkelMo shows consistent performance with only minor degradation under auto-rigged rigs.}
\vspace{-0.2cm}
  \label{tab:autorig}
  \centering
  \begin{tabular}{@{}l|c|c|c|c|c@{}}
    \toprule
    Metrics& Motion& Puppeteer& ActionMesh& SkelMo & SkelMo\\
    $(\times 10^{-1})$& 3-to-4 & & & (w. GT) & (w. Auto-rigged) \\
    \midrule
    MSE$\downarrow$ & 0.369 & 0.229 & 0.265 & \textbf{0.048} & 0.082\\
    Vertex-CD$\downarrow$ & 0.230 & 0.158 & 0.173 & \textbf{0.107} & 0.146\\
  \bottomrule
  \end{tabular}
 
 \vspace{-7pt}
\end{table}

\subsection{Application}
\begin{figure}[t]
    \centering
    \includegraphics[width=\linewidth]{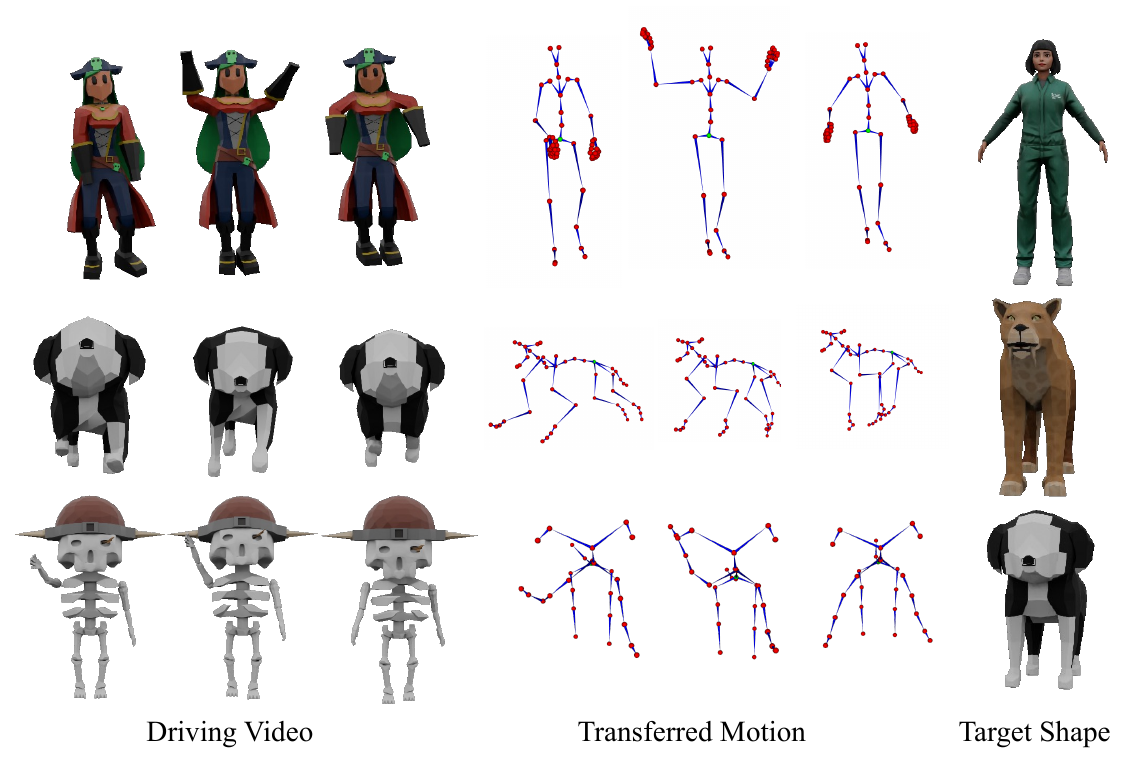}
    \vspace{-0.7cm}
    \caption{Visualization of cross-category motion transfer results. Each row represents a specific transfer case across multiple frames. The leftmost column displays the target static meshes, while the subsequent columns show pairs of the driving video (source motion) and the resulting target skeleton. The first and second rows demonstrate intra-category transfer (human-to-human and quadruped-to-quadruped, respectively), while the third row illustrates cross-species transfer (e.g., human-to-quadruped). The extracted skeletons validate the model’s robustness in retargeting complex motions across significantly different skeletal morphologies and joint configurations.
  }
  
  \label{fig:transfer}
\end{figure}

\noindent\textbf{Motion Transfer.} Beyond standard motion generation, the topology-agnostic nature of SkelMo enables the challenging task of cross-entity motion transfer. 
In this setting, we replace the source video with a driving sequence featuring a completely different body shape or species (e.g., transferring motion from a quadruped to a humanoid, or vice versa).
As illustrated in Fig.~\ref{fig:transfer}, our model successfully preserves the core motion dynamics of the driving video while faithfully adapting them to the geometric constraints of the target skeleton. 
The first two rows of Fig.~\ref{fig:transfer} demonstrate intra-category transfer (human-to-human and quadruped-to-quadruped), where the model accurately captures fine-grained skeletal movements. 
More importantly, the third row highlights the capability of the model in cross-species transfer (e.g., human-to-quadruped skeleton).

This robust adaptation stems from our representation using global Cartesian coordinates, which provides a universal geometric language for motion, and the Skeletal Attention mechanism, which re-targets these coordinates to remain structurally valid for the new topology.
Even with significant discrepancies in bone proportions and joint hierarchies, SkelMo produces natural and synchronized animations, highlighting its robustness in complex cross-category motion synthesis.

\begin{figure}[t]
    \centering
     \includegraphics[width=\linewidth]{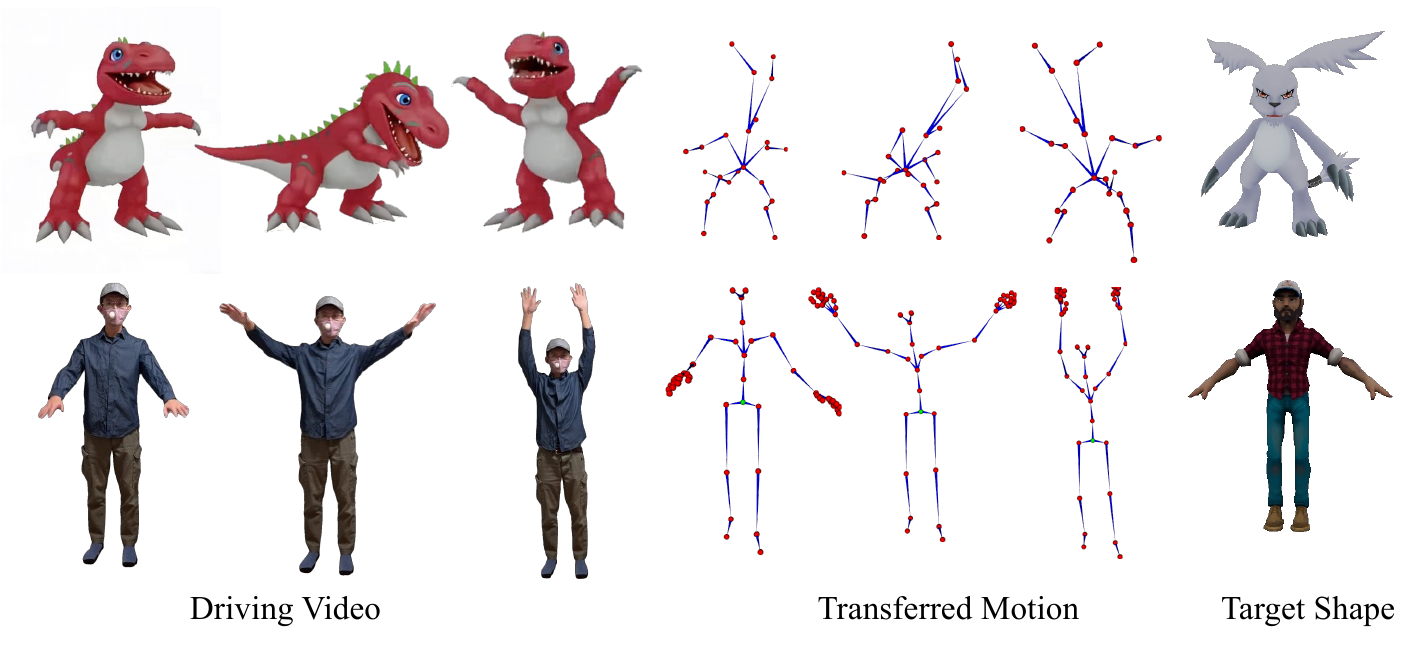}
     \vspace{-0.7cm}
    \caption{Visualization of in-the-wild scenarios. Our method demonstrates strong generalization across diverse driving video sources. The first row showcases results driven by AI-generated videos (e.g., from Doubao AI), featuring stylized motions. The second row illustrates performance on real-world captures~\cite{yang2022banmo}, highlighting the stability of the model under complex lighting and backgrounds. In both cases, high-fidelity motion retargeting is consistently achieved.
  }
  \vspace{-0.4cm}
  \label{fig:real}
\end{figure}

\vspace{0.5em}
\noindent\textbf{In-the-Wild Scenarios.} To further evaluate the robustness and practical applicability of our method, we conduct experiments using driving videos from two distinct sources: AI-generated videos and real-world captures.
Specifically, we utilize high-quality synthetic videos generated by AI (e.g., Doubao AI~\cite{doubao2026}) as driving signals to test the model's performance on stylized motions. 
The first row of Fig.~\ref{fig:real} demonstrates this by successfully transferring motion from an AI-generated character to a different target shape.
Additionally, we apply our method to raw in-the-wild real-world videos to assess its stability under complex lighting and diverse backgrounds. 
The second row illustrates this performance on real-world captures where high-fidelity motion retargeting is consistently achieved. 
These results across both scenarios demonstrate strong generalization capabilities for various downstream applications.

\section{Conclusion and Discussion}
In this paper, we presented SkelMo, a diffusion-based framework designed to bridge the gap between 2D visual motion cues and 3D skeletal animation for category-agnostic rigged objects. 
By curating a large-scale, high-fidelity dataset of approximately 20,000 3D aimations with complete rigging and diverse animations, we have addressed a critical data bottleneck in the field. 
By leveraging a bidirectional video-skeleton fusion and a texture-semantic injection mechanism, our architecture enables the model to transcend the limitations of species-specific templates, achieving robust and physically plausible motion synthesis across diverse, unseen morphologies.

Despite these advancements, several promising directions remain for future exploration. 
A primary goal is to enrich our dataset with fine-grained textual captions that describe the stylistic nuances and intent of each motion sequence. 
Such semantic grounding will facilitate the transition from video-only conditioning to multimodal generative control, enabling users to synthesize complex animations through text instructions or audio signals. 
Furthermore, exploring object-environment interactions and physically-aware constraints will be crucial for extending the capabilities of SkelMo toward more complex embodied AI tasks. 
By expanding both the semantic depth of our data and the flexibility of our conditioning modalities, we aim to provide a truly universal engine for bringing diverse virtual and physical entities to life.

\section*{Acknowledgements}
This work was supported in part by the Hong Kong Research Grants Council under Grants R1018-25F, N\_CityU1114/25, and 11219324; and in part by the National Natural Science Foundation of China under Grant 62422118.
\clearpage
\bibliographystyle{splncs04}
\bibliography{main}
\end{document}